%% file: main.tex
\documentclass[letterpaper, 10 pt, conference]{ieeeconf}  

\makeatletter

\newcommand{\Rmnum}[1]{\expandafter\@slowromancap\romannumeral #1@}
\makeatother

\IEEEoverridecommandlockouts                              

\usepackage[utf8]{inputenc} 
\usepackage[T1]{fontenc}    
\usepackage{hyperref}       
\usepackage{url}            
\usepackage{booktabs}       
\usepackage{amsfonts}       
\usepackage{amsmath}
\usepackage{amssymb}
\usepackage{nicefrac}       
\usepackage{microtype}      
\usepackage{graphicx}
\usepackage{comment}
\usepackage{multirow}
\usepackage{bm}
\usepackage{mathtools}
\usepackage{multirow}
\usepackage{cite} 
\usepackage{subfigure}

\title{\LARGE \bf
Sequence-to-Sequence Prediction of Vehicle Trajectory\\via LSTM Encoder-Decoder Architecture
}

\author{Seong Hyeon Park, ByeongDo Kim, Chang Mook Kang, Chung Choo Chung and Jun Won Choi$^{*}$ \\ Hanyang University
\thanks{$^{*}$ corresponding author}%
\thanks{The authors are with the Department of Electrical Engineering, Hanyang\newline University, Seoul, Korea.
        \newline{\tt\small \{shpark,bdkim\}@spo.hanyang.ac.kr\newline \{kcm0728,cchung,junwchoi\}@hanyang.ac.kr}}%
}

\begin{document}

\maketitle
\thispagestyle{empty}
\pagestyle{empty}

\begin{abstract}
  \input{./tex/abstract}
\end{abstract}

\section{Introduction}
	\input{./tex/intro}

\section{Review on LSTM Encoder-Decoder Architecture}
	\input{./tex/review}

\section{Proposed Trajectory Analysis Technique} 
	\input{./tex/proposed}

\section{Experiments} 
	\input{./tex/exp}
    
\section{Conclusions}
	\input{./tex/conclusion}




\section*{ACKNOWLEDGMENT}
This work was supported by the Technology Innovation Program(10083646) funded By the Ministry of Trade, Industry \& Energy(MOTIE, Korea) and Institute for Information \& communications Technology Promotion(IITP) grant funded by the Korea government(MSIT) (No. R7117-16-0164, Development of wide area driving environment awareness and cooperative driving technology which are based on V2X wireless communication)
\bibliography{bibtex/bib/IEEEabrv.bib,bibtex/bib/simon.bib}{}
\bibliographystyle{IEEEtran}
\end{document}

%% file: tex/abstract.tex
In this paper, we propose a deep learning based vehicle trajectory prediction technique which can generate the future trajectory sequence of surrounding vehicles in real time. We employ the encoder-decoder architecture which analyzes the pattern underlying in the past trajectory using the long short-term memory (LSTM) based encoder and generates the future trajectory sequence using the LSTM based decoder. This structure produces the $K$ most likely trajectory candidates over occupancy grid map by employing the {\it beam search} technique which keeps the $K$ locally best candidates from the decoder output. The experiments conducted on highway traffic scenarios show that the prediction accuracy of the proposed method is significantly higher than the conventional trajectory prediction techniques.


%% file: tex/intro.tex
  Ensuring safety is a top priority for the autonomous driving and advanced driver assistance systems (ADAS). In order to promise high degree of safety, the ability to perceive surrounding situations and predict their development in the future is critical. The vehicle on driving encounters various types of dynamic traffic participants such as car, motor bike, and pedestrian which could be a potential threat to safe driving. In order to avoid an accident, the system should be able to analyze the pattern in their motion and predict the future trajectories in advance. If the system predicts where the surrounding vehicles are heading in the near future, the vehicle can plan its driving path in response to the situation to come such that  the probability of collision is minimized. However, the trajectory of the surrounding vehicles is quite complex to analyze since it is governed by various latent factors determined by complex traffic situations and the state of these latent factors can change dynamically in real time. 

  
  Thus far, various vehicle trajectory analysis techniques have been proposed. The traditional approaches are the Bayesian filtering methods such as Kalman and extended Kalman filters \cite{ammoun2009real,barrios2011improving}. However, the structure of these methods might be too simple to analyze the complicated pattern of the vehicle motion and they do not often perform well for long term prediction, e.g., $\Delta = 2$ sec. In order to overcome the limitation, more sophisticated trajectory models were introduced, including Gaussian process model \cite{tran2014online,laugier2011probabilistic}, Gaussian mixture model \cite{wiest2012probabilistic}, and dynamic Bayesian network (DBN) \cite{gindele2015learning}. In particular, the DBN provides a flexible trajectory analysis framework where various latent factors determining the vehicle trajectory are described using the graphical model, and the interactions among these factors are learnt from the data. Though the DBN leads to explicit modeling of physical process that generates a vehicle's trajectory, the performance for real traffic scenarios is limited in that the model structure determined by the designer's intuition is not sufficient to capture a variety of dynamic traffic scenarios. Furthermore, the computational complexity for its inference step is quite high for the real time applications.
  
\begin{figure}[t]
	\centering
	\includegraphics[width=82mm]{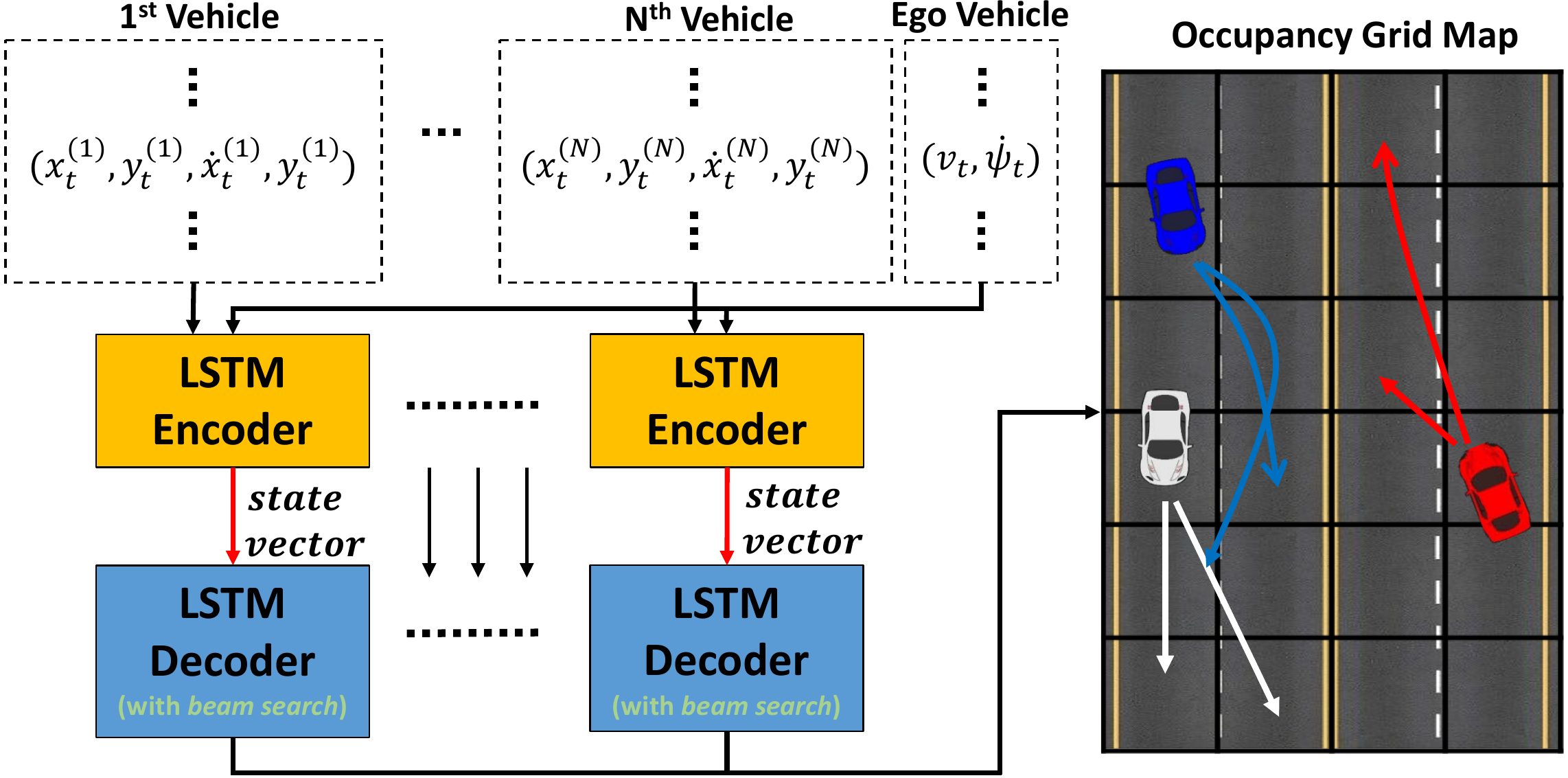}
	\caption{The proposed trajectory prediction system. $(x^{(n)}_{t},y^{(n)}_{t},\dot{x}^{(n)}_{t},\dot{y}^{(n)}_{t})$ denotes the $n$th surrounding vehicle's relative position and velocity at time $t$ and $(v_{t},\dot{\psi}_{t})$ denotes the ego vehicle's speed and yaw rate at time $t$.}
	\label{fig:modelstructure}
\end{figure}

  Meanwhile, the era of deep learning has begun with the success of many revolutionary deep neural network (DNN) architectures \cite{krizhevsky2012imagenet, cho2014learning, goodfellow2016deep}. Being applied to numerous machine learning tasks, DNN architectures have successfully learned the representation that generalizes well for various situations arising in the real data. Among various kinds of DNN architectures available, recurrent neural network (RNN) is widely used to analyze the structure of the time series data. One popular variant of RNN is {\it long short-term memory} (LSTM) model, which can control the information flow between the input, output, and cell memory through the gating mechanism \cite{hochreiter1997long}. The LSTM has shown excellent performance for various tasks such as speech recognition, image captioning, language translation and so on \cite{karpathy2015rnn}. Recently, LSTM or similar RNN variants have also been applied to analyze the vehicle trajectory \cite{ondruska2016deep,khosroshahi2016surround,kim2017probabilistic,lee2017desire}. In \cite{ondruska2016deep}, the LSTM is used to track the position of the object based on the ranging sensor measurements. In \cite{khosroshahi2016surround}, the driver's intention is identified based on the trajectory data using the LSTM. In \cite{kim2017probabilistic}, the LSTM is applied to predict the location of the vehicle after $\Delta$ seconds using the past trajectory data. In \cite{lee2017desire}, another RNN variant called gated recurrent unit (GRU) combined with conditional variational auto-encoder (CVAE) is used to predict the vehicle trajectory. Although these RNN based vehicle trajectory models are effective for their own tasks, they either cannot generate the full trajectory sequence or do not provide a simple probabilistic framework that does not require extra components such as CVAE.
  
  In this paper, we propose a new vehicle trajectory analysis and prediction technique to generate the future trajectory of surrounding vehicles given the sequence of the latest sensor measurements. The proposed method is built upon the LSTM encoder-decoder architecture which has shown excellent performance for {\it sequence to sequence} tasks \cite{cho2014learning}. Fig.~\ref{fig:modelstructure} depicts the structure of the proposed technique. The LSTM encoder takes the latest trajectory samples for the surrounding vehicles as well as the state information on the ego vehicle and produces the fixed length vector which captures the temporal structure of the past trajectory. Based on the fixed length vector, the LSTM decoder generates the future trajectory on the occupancy grid map (OGM). The decoder recursively uses the predicted trajectory sample to generate the subsequent trajectory samples adopting {\it beam search} algorithm. With the algorithm, the decoder keeps $K$ locally best sequence candidates in generating the future trajectory sample for each time step \cite{neubig2017neural}. As a result, the proposed model can predict the $K$ most probable hypotheses of the vehicle trajectory under the probabilistic framework. Our experiment results show that the proposed scheme generates the reasonable trajectory hypotheses of the surrounding vehicles and its prediction accuracy is significantly better than the conventional prediction methods.

  The rest of this paper is organized as follows. In Section \Rmnum{2}, we briefly review the LSTM encoder-decoder architecture and the beam search algorithm. In Section \Rmnum{3}, we describe the proposed trajectory prediction model in details. In Section \Rmnum{4}, the experimental results are presented and Section \Rmnum{5} concludes this paper.

%% file: tex/review.tex
\subsection{Long-Short Term Memory  (LSTM)}
The long short-term memory (LSTM) is an RNN variant which effectively overcomes the vanishing gradient issue in naively designed RNNs \cite{hochreiter1997long}. The LSTM consists of the cell memory that stores the summary of the past input sequence, and the gating mechanism by which the information flow between the input, output, and cell memory are controlled. The following recursive equations describe how the LSTM works;
\begin{align}
	\bm{f}_t =~&\sigma(W_{uf} \bm{u}_t + W_{hf}\bm{h}_{t-1} + \bm{b}_f) \label{eqn:forgetgate}  \\
    \bm{i}_t =~&\sigma(W_{ui} \bm{u}_t + W_{hi}\bm{h}_{t-1} +\bm{b}_i) \label{eqn:inputgate}  \\
	\bm{o}_t =~&\sigma(W_{uo} \bm{u}_t + W_{ho}\bm{h}_{t-1} + \bm{b}_o) \label{eqn:outputgate} \\
    \begin{split}
	\bm{c}_t =~&\bm{f}_{t} \odot \bm{c}_{t-1} \\
    &+ \bm{i}_t \odot \tanh(W_{uc} \bm{u}_t + W_{hc}\bm{h}_{t-1} + \bm{b}_c) \label{eqn:cell} \end{split}\\
    \bm{h}_t =~&\bm{o}_t \odot \tanh(\bm{c}_{t}), \label{eqn:output}
\end{align}
where
\begin{itemize}
    \item $\sigma({x}) \triangleq \frac{1}{1+\exp{(-x)}}$: sigmoid function (element-wise)
	\item $x \odot y$: element wise product
    \item $\bm{u}_t$: input vector
	\item $W_{ui}, W_{hi}, W_{uf}, W_{hf}, W_{uo}, W_{ho},W_{uc},W_{hc}$:\newline linear transformation matrices
	\item $\bm{b}_i, \bm{b}_f, \bm{b}_o, \bm{b}_c$: bias vectors
    \item $\bm{i}_t, \bm{f}_t, \bm{o}_t$: gating vectors
	\item $\bm{c}_t$: cell memory state vector
    \item $\bm{h}_t$: state output vector.
\end{itemize}
The amount of information for the cell memory to update, forget, and output its state is determined by the gating vectors in \eqref{eqn:forgetgate}, \eqref{eqn:inputgate}, and \eqref{eqn:outputgate}. The cell state and output are updated according to \eqref{eqn:cell} and \eqref{eqn:output}. Note that depending on the state of the forget gating vector $\bm{f}_t$, the cell state can be reset or restored, and the other two gating vectors $\bm{i}_t$ and $\bm{o}_t$ operate in the similar manner to regulate the input and output.

\subsection{LSTM Encoder-Decoder Architecture}
\begin{figure}[t]
	\centering
    \vspace{0.16cm}
	\includegraphics[width=85mm]{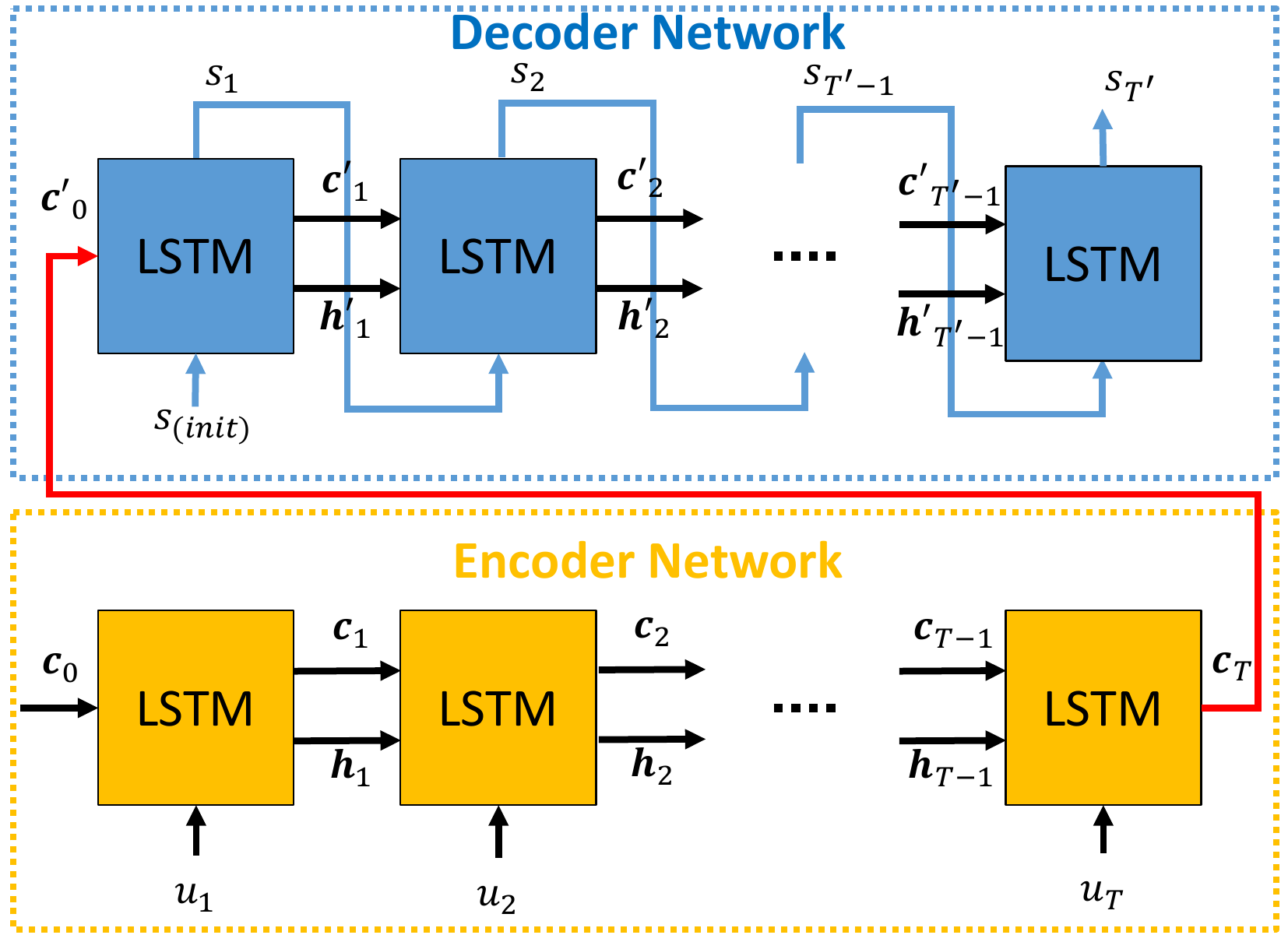}
	\caption{The LSTM encoder-decoder architecture.}
	\label{fig:encdec}
\end{figure}
The LSTM encoder-decoder architecture was first introduced for machine translation task \cite{cho2014learning,neubig2017neural,luong17}. It has an ability to read and generate a sequence of arbitrary length as illustrated in Fig.~\ref{fig:encdec}. The architecture employs two LSTM networks called the encoder and decoder. The encoder processes the input sequence ${u}_1,...,{u}_T$ of the length $T$ and produces the summary of the past input sequence through the cell state vector $\bm{c}_t$. After $T$ times of recursive updates from \eqref{eqn:forgetgate} through \eqref{eqn:output}, the encoder summarizes the whole input sequence into the final cell state vector $\bm{c}_T$. Then, the encoder passes $\bm{c}_T$ to the decoder so that the decoder uses it as initial cell state (i.e., $\bm{c'}_0 = \bm{c}_T$) for the sequence generation. The decoding step is initiated with a dummy input ${s}_{\text{(init)}}$. The decoder recursively generates the output sequence ${s}_{1},...,{s}_{T'}$ of the length $T'$. In every update, the decoder feeds the output $s_{t-1}$ obtained in the previous update to the input for the current update. Note that the output of the decoder are derived by applying the affine transformation followed by the function that suits for the specific tasks (e.g. Softmax function for classification task). 

  Basically, the LSTM encoder-decoder aims to model the conditional probability of the output sequence given the input sequence, i.e., $p({s}_{1},...,{s}_{T'}|{u}_1,...,{u}_T)$. The encoder provides the summary of the input sequence ${u}_1,...,{u}_T$ through the LSTM cell state $\bm{c}_T$. Given the encoder cell state $\bm{c}_T$, the conditional probability is approximated to
\begin{align}
	p({s}_{1},...,{s}_{T'}|{u}_1,...,{u}_T)   \label{eqn:condprob}
    &\approx \prod_{t=1}^{T'}p({s}_{t}|\bm{c}_T,{s}_{1},...,{s}_{t-1}). 
\end{align}
The decoder successively produces the probability distribution of $p({s}_{t}|\bm{c}'_{t-1},{s}_{t-1})$ given the decoder cell state $\bm{c}'_{t-1}$ and the $(t-1)$th sample of the output sequence $s_{t-1}$, i.e.,  
\begin{align}
	p({s}_{1},...,{s}_{T'}|{u}_1,...,{u}_T)  \approx \prod_{t=1}^{T'}p({s}_{t}|\bm{c}'_{t-1},{s}_{t-1}).
\end{align}
Unfortunately, the decoder does not know the true value of the previous output sample. Hence, in every decoding step, the decoder makes a decision on ${s}_{t}$ based on the probability distribution $p({s}_{t}|\bm{c}'_{t-1},{s}_{t-1})$ obtained from the decoder output and use the tentative decision for the next update of the decoder state. 

\subsection{The Beam-Search Algorithm}
As mentioned, the LSTM decoder aims to produce the probability distribution of ${s}_{t}$ given the decoder cell state $\bm{c}'_{t-1}$ and the $(t-1)$th output sample $s_{t-1}$. One way to determine $s_{t}$ is the greedy search strategy that simply picks the value for ${s_{t}}$ that maximizes the probability $p({s_{t}}|\bm{c}'_{t-1},s_{t-1})$ and feed it back to the decoder to generate the next output sample. Unfortunately, such greedy strategy suffers from the error propagation since wrong decision made at the current time step would be propagated to the subsequent time steps. Furthermore, $p({s_{t}}|\bm{c}'_{t-1},s_{t-1})$ might be a multi modal distribution having multiple peaks which correspond to the equally promising candidates for ${s}_{t}$. In this case, generating only a single hypothesis is not sufficient to represent all probable outcomes.

  In order to alleviate the error propagation, the {\it beam search} algorithm has been introduced for the machine translation tasks \cite{neubig2017neural}. The basic idea of the beam search is to keep the $K$ most probable hypotheses for each sequence generation step where the parameter $K$ is called {\it beam width}. For each of $K$ hypothetical sequences found in the previous decoding step, the decoder generates $|S|$ candidates for $s_t$ resulting in total $K\times|S|$ candidates where $|S|$ denotes the cardinality of the set of all possible selections for $s_{t}$. Then, it chooses the $K$ best hypotheses according to the conditional probability $p({s_{t}}|\bm{c}'_{t-1},s_{t-1})$ produced at the decoder output. After $T'$ iterations, the $K$ best hypotheses would survive as a final result of the decoding step. Note that the beam search with $K=1$ reduces to the greedy search algorithm.

%% file: tex/proposed.tex
  In this section, we describe the system design, the detailed network structure, and the training methodology of the proposed technique.

\subsection{System Description}
\begin{figure}[tbh]
	\centering
    \begin{subfigure}[]
		{\includegraphics[width=2.7in]{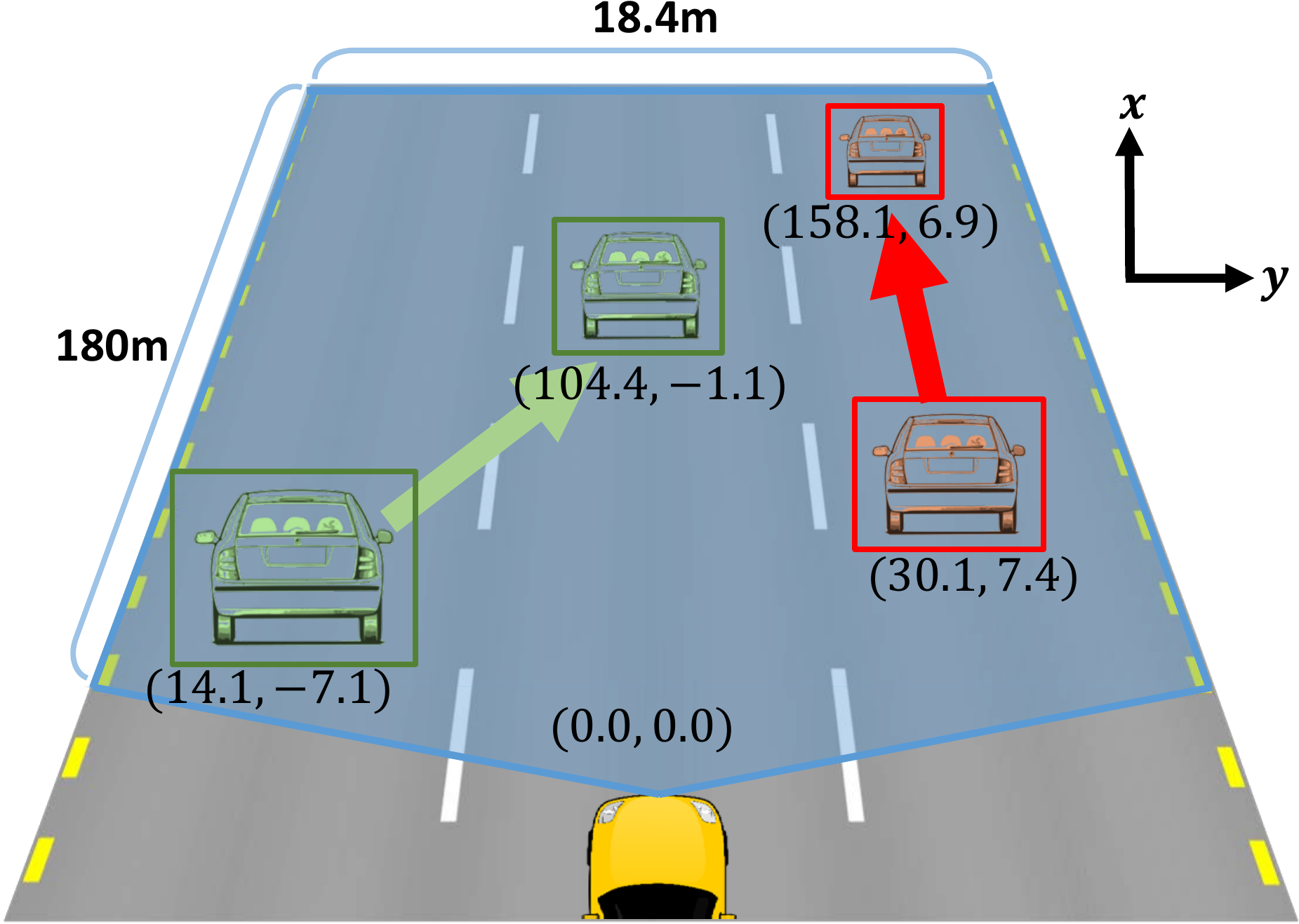}}
    \end{subfigure}
    \begin{subfigure}[]
    	{\includegraphics[width=2.7in]{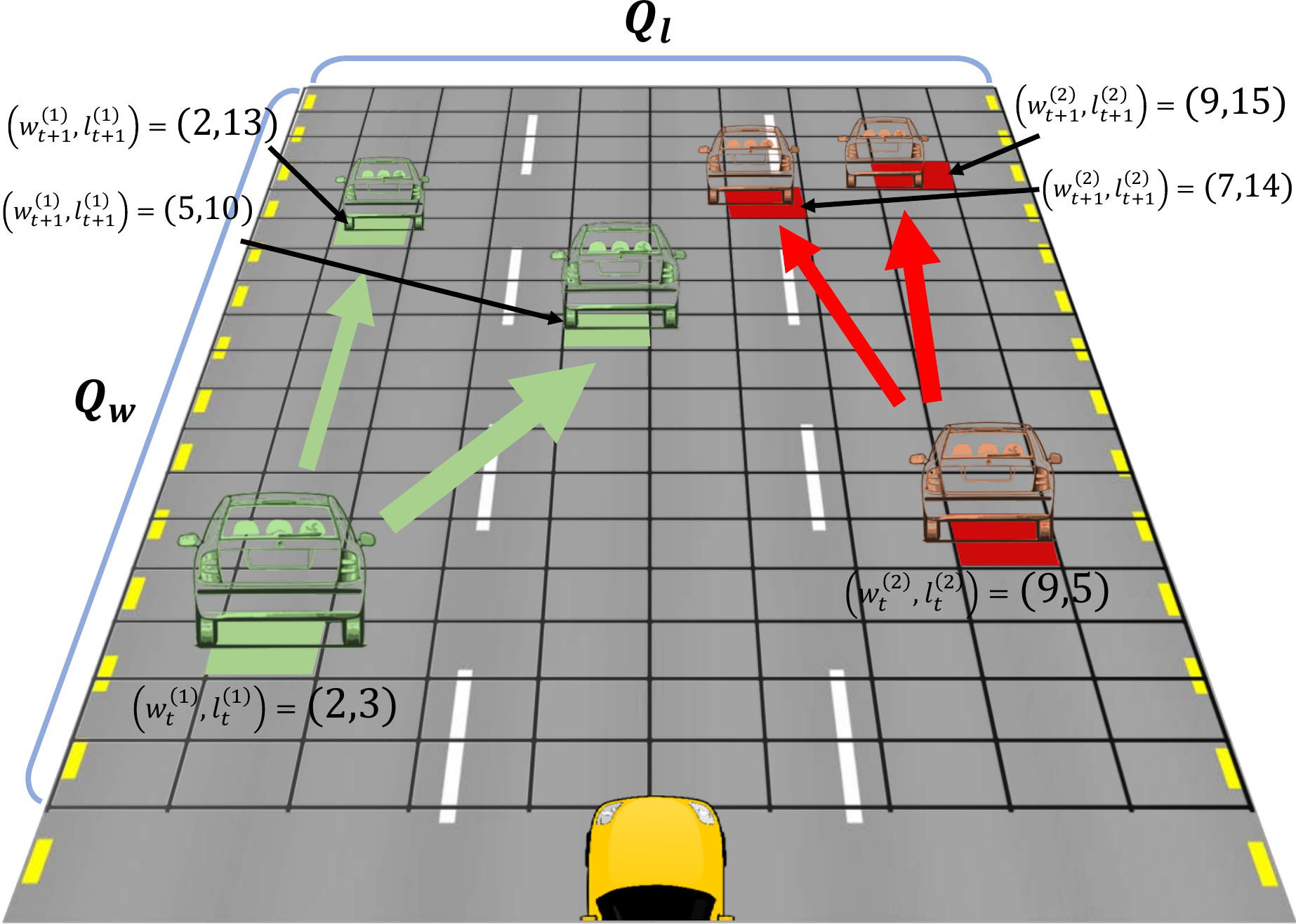}}
    \end{subfigure}
	\caption{The description of (a)  relative coordinate system and (b) grid representation on occupancy grid map.}
	\label{fig:rel_coord}
\end{figure}

\begin{figure*}[t]
\centering
\vspace{0.16cm}
\includegraphics[width=15cm]{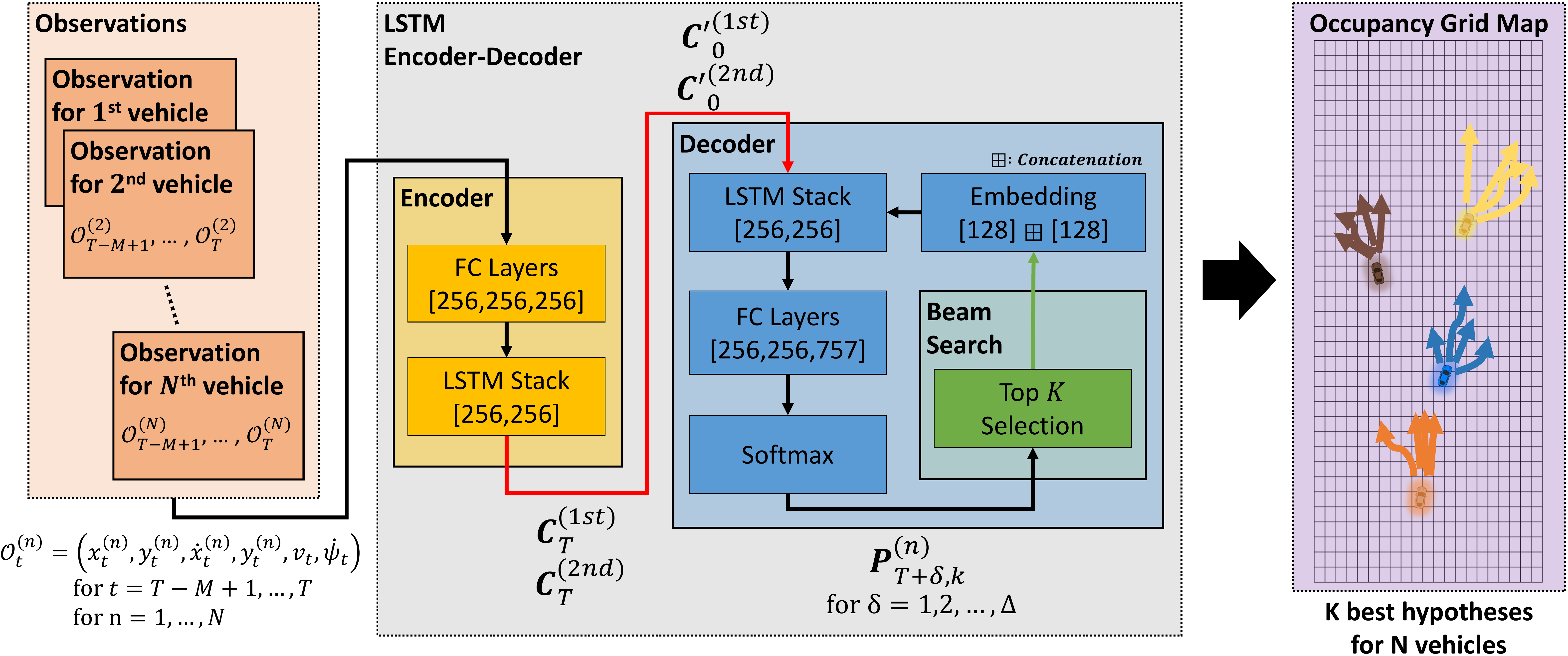}
\caption{The overall structure of the proposed trajectory analysis technique.}
\label{fig:detailedstructure}
\end{figure*}

  Our system assumes that the ego vehicle can estimate the current state of the relative coordinate and velocity of the surrounding vehicles from the measurements acquired by the sensors (e.g. camera, lidar, and radar sensors). In addition, it can obtain the information on its own motion using the inertial measurement unit (IMU) sensor. The set of the observations used for prediction of the trajectory of the $n$th surrounding vehicle at the time step $t$ is given by 
\begin{align}\mathcal{O}_t^{(n)} = \{{v}_{t},{\dot{\psi}_{t}},{x}^{(n)}_{t},{y}^{(n)}_{t},{\dot{x}}^{(n)}_{t},{\dot{y}}^{(n)}_{t}\}
\end{align}
where
\begin{itemize}
	\item ${v_{t}}$: the ego vehicle's speed
	\item ${\dot{\psi}_{t}}$: the ego vehicle's yaw rate 
    \item $\{{x^{(n)}_{t}},{y^{(n)}_{t}}\}$:  the $n$th vehicle's relative coordinate 
    \item $\{{\dot{x}^{(n)}_{t}},{\dot{y}^{(n)}_{t}}\}$: the $n$th vehicle's relative velocity.
\end{itemize}
As illustrated in Fig.~\ref{fig:rel_coord} (a), we adopt the relative coordinate system where the ego vehicle's location is fixed to $(0,0)$. The longitudinal and lateral ranges are set to be $x\in[0, 180]$ and $y\in[-9.2, 9.2]$ in meters which are determined according to the valid detection range of the sensors.

  In order to represent the future trajectory of the surrounding vehicles in the proposed system, we use the occupancy grid map (OGM) which has been widely used in robotics for the object localization \cite{milstein2008occupancy}. The OGM divides the region around the ego vehicle into $Q_w \times Q_l$ rectangular grid elements. As shown in Fig.~\ref{fig:rel_coord} (b), we represent each element of OGM by the grid index. In the proposed system, we use $(36 \times 21)$ grids, where each grid span $5.0$ meter by $0.875$ meter region in longitudinal and lateral directions, respectively. This design is adopted in order for a single grid to cover the length of a typical sedan and the quarter lane width. Using the OGM, the trajectory prediction task becomes a classification problem with $Q = Q_w \times Q_l +1$ classes, each of which corresponds to one of the $Q_w \times Q_l$ grid elements on the OGM or the out-of-map state. We formulate the vehicle trajectory prediction as a sequential multiclass classification problem where the grid element occupied by a surrounding vehicle should be chosen sequentially for each time step.

  The overall workflow of the system is illustrated in Fig.~\ref{fig:detailedstructure} and described as follows. For the $n$th surrounding vehicle, the sequence of the latest $M$ observations $\mathcal{O}_{T-M+1}^{(n)}, ..., \mathcal{O}_{T}^{(n)}$ is fed into the encoder. Given the latest cell memory produced by the encoder through $M$ updates, the decoder sequentially generates the future trajectory for $\Delta$ time steps ahead. Note that using the beam search, the decoder generates the $K$ most likely trajectory sequences in parallel. At the $(T+\delta)$th time step ($\delta \in \{1, ..., \Delta\}$), the decoder calculates the probability $P_{T+\delta,k,q}^{(n)}$ that the $n$th vehicle occupies the OGM element indexed by $q\in\{1,...,Q\}$.
   This constructs the probability map of size $Q$ i.e., $\bm{P}_{T+\delta,k}^{(n)} = \{{P}_{T+\delta,k,1}^{(n)},..., {P}_{T+\delta,k,Q}^{(n)} \}$. Since this decoding process is performed for all survived $K$ trajectory sequences, we come up with $K$ probability maps $\bm{P}_{T+\delta,1}^{(n)}, ..., \bm{P}_{T+\delta,K}^{(n)}$. Then, based on the $K$ probability maps, we pick the $K$ best trajectory candidates which yield the largest probability of occupancy. Note that these candidates are fed back to the LSTM decoder through the data embedding process. Repeating the above process with $\Delta$ updates, the decoder ends up with the final $K$ best trajectory sequences for the $n$th surrounding vehicle. Sharing the parameters of the encoder and decoder for all surrounding vehicles,
we can collect the results of trajectory prediction for $N$ surrounding vehicles on the single OGM. A total of $K\times N$ future trajectory hypotheses shown on the single OGM presents the unified view on how the state of the dynamic objects around the ego vehicle would develop in $\Delta$ time steps ahead. 

\subsection{Network Structure}

\subsubsection{Encoder}
The encoder consists of three fully connected (FC) layers followed by two LSTM layers stacked. Each FC layer contains affine transformation followed by ReLU (Rectified Linear Unit) activation function. They are designed for two purposes; 1) transforming the $6$ dimensional input data into $256$ dimensional feature being consistent with the LSTM cell dimension and 2) extending the network capacity enough to capture the complex structure of the trajectory data.  The output from the last FC layer is then fed into the LSTM stack. The LSTM stack is constructed with two LSTMs with $256$ dimensional cell memory for each. Inside the stack, the output vector from the the first LSTM is fed to the second LSTM as an input. After $M$ recursive updates in the two LSTMs, their latest cell states $\bm{C}^{\text{(1st)}}_{T}$ and $\bm{C}^{\text{(2nd)}}_{T}$ are determined and passed to the decoder.

\subsubsection{Decoder}
The decoder consists of the two LSTM layers stacked followed by three FC layers, the Softmax layer, and the embedding layer. The decoder LSTMs use the cell state vectors passed from the encoder as their initial cell states. The first LSTM of the decoder is initialized by $\bm{C}^{\text{(1st)}}_{T}$ and the second decoder LSTM is by $\bm{C}^{\text{(2nd)}}_{T}$. The output from the second LSTM is fed to the FC layers. While the first and second FC layers have the same structure as those in the encoder, the last FC layer has the output dimension of $757$ corresponding to the number of class (i.e., $Q=36\times21+1$). Then, the output of the FC layers is passed through the Softmax function, which produces the probability of the occupancy on the probability map of the size 757. Based on all $K$ probability maps derived for all hypothesis, most probable $K$ trajectory candidates are selected by the beam search algorithm. The selected $K$ OGM indices are fed back to the decoder through the embedding process. In the embedding step, we employ two embedding matrices of size $128\times37$ and $128\times22$ in which each column vector corresponds to the longitudinal (${w}_{t}^{i} \in \{1,...,36\}$) and lateral (${l}_{t}^{i} \in \{1,...,21\}$) grid indices on the OGM as well as the out-of-map state. Among these column vectors, $K$ vectors are gathered from each embedding matrix, that is, $2 \times K$ vectors in total according to the selected longitudinal and lateral OGM indices. Then, they are concatenated into $256$ dimensional vectors where each of them represents the selected OGM indices and delivered to the decoder LSTM to be used in the next decoding step. This procedure is repeated until we reach the prediction length of $\Delta$.

\subsection{Training Methodology}
The parameters of the LSTM encoder-decoder (including the embedding matrices) are trained in an end to end fashion. We generate the training data set by cropping all available $(M+\Delta)$ length trajectory samples from the trajectory record for each surrounding vehicles. Assuming that the trajectory data collected from the sensors $\mathcal{O}_{T-M+1}^{(n)}, ..., \mathcal{O}_{T+\Delta}^{(n)}$ is sufficiently good in terms of accuracy, the OGM grid indices extracted from the subsequent $\Delta$ measurements (i.e., $\mathcal{O}_{T+1}^{(n)}, ..., \mathcal{O}_{T+\Delta}^{(n)}$) can be used as the label for the supervised training. The loss function to be minimized is given by the negative log likelihood function 
\begin{align}
\begin{split}
   	L(\theta) =& -\sum_{j=1}^{J} \sum_{\delta=1}^{\Delta} \sum_{q=1}^{Q}\Big( o_{j,T+\delta,q} \ln z_{j,T+\delta,q}\Big)
\end{split}
\end{align}
where  $J$ is the total number of the training samples, $\Delta$ is the prediction length, and $Q$ is the total number of classes. The variable $o_{j,T+\delta,q}$ is the OGM grid label represented in the one hot encoding.  For the $j$th example,  the label $o_{j,T+\delta,q}$ becomes one if the surrounding vehicle occupies the $q$th grid element and zero otherwise. Note also that $z_{j,T+\delta,q}$ is the $q$th output of the Softmax layer in the LSTM decoder for the $j$th example. For the minimization of the loss function, we adopt the stochastic gradient decent method with a momentum, called {\it ADAM} optimizer \cite{kingma2014adam} with mini batch size $B$. The training is stopped if the validation error (obtained from 15\% of the training data) does not decrease anymore.

%% file: tex/exp.tex
In this section, we present the performance of the proposed trajectory prediction technique based on the experiments conducted on real highway driving.
\begin{figure}[t]
	\centering
    \vspace{0.16cm}
	\includegraphics[width=82mm]{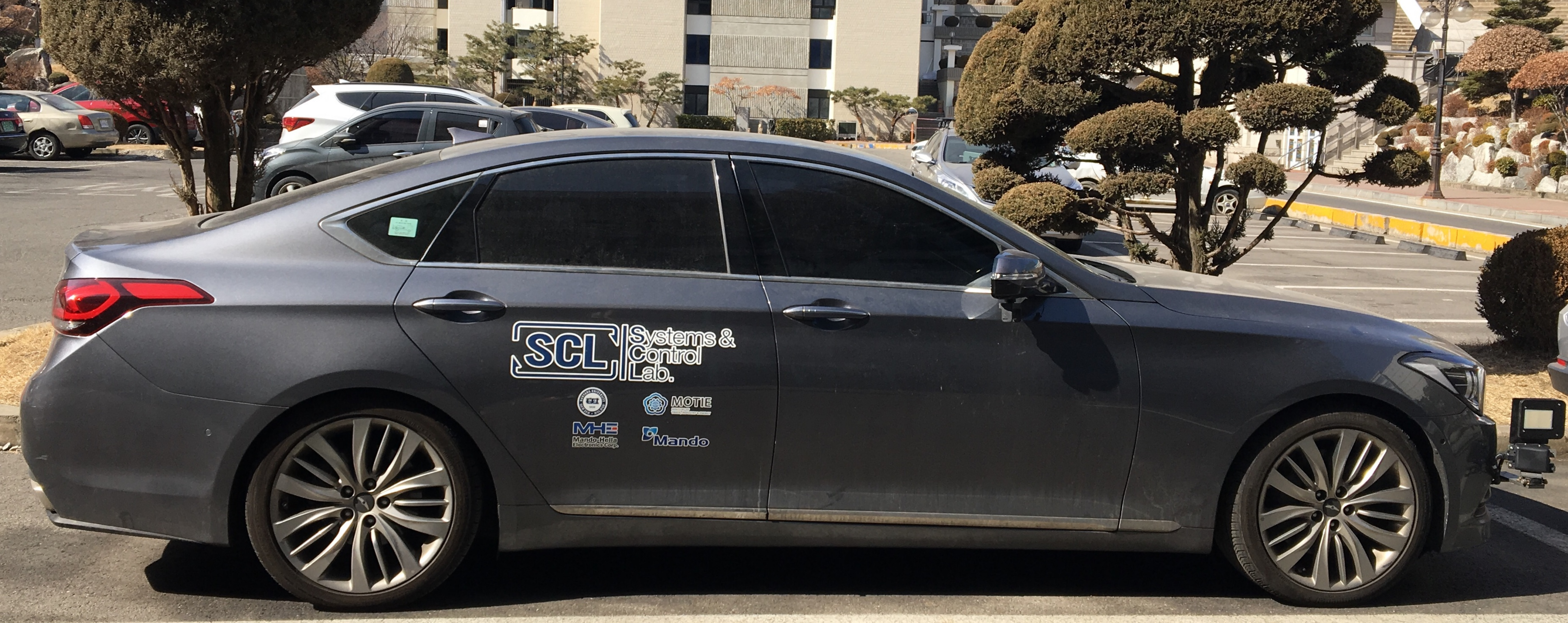}
	\caption{The test vehicle.}
	\label{fig:genesis}
\end{figure}

\begin{table*}[t]
\centering
\vspace{0.16cm}
\caption{MAE, MAE\_X, and MAE\_Y of several trajectory prediction algorithms}
\label{table:mae}
\begin{tabular}{ccccccc}
\multicolumn{1}{l}{} & \multicolumn{6}{c}{\textbf{MAE (Grids)}} \\ \cline{2-7} 
\multicolumn{1}{l|}{} & \multicolumn{1}{c|}{\multirow{2}{*}{\textbf{\begin{tabular}[c]{@{}c@{}}Proposed\\ ($\Omega = 1$)\end{tabular}}}} & \multicolumn{1}{c|}{\multirow{2}{*}{\textbf{\begin{tabular}[c]{@{}c@{}}Proposed\\ ($\Omega = 3$)\end{tabular}}}} & \multicolumn{1}{c|}{\multirow{2}{*}{\textbf{\begin{tabular}[c]{@{}c@{}}Proposed\\ ($\Omega = 5$)\end{tabular}}}} & \multicolumn{1}{c|}{\multirow{2}{*}{\textbf{\begin{tabular}[c]{@{}c@{}}Baseline 1\\ (Kalman filter)\end{tabular}}}} & \multicolumn{1}{c|}{\multirow{2}{*}{\textbf{\begin{tabular}[c]{@{}c@{}}Baseline 2\\ (Basic LSTM)\end{tabular}}}} & \multicolumn{1}{c|}{\multirow{2}{*}{\textbf{\begin{tabular}[c]{@{}c@{}}Baseline 3\\ \cite{kim2017probabilistic}\end{tabular}}}} \\ \cline{1-1}
\multicolumn{1}{|c|}{\textbf{$\Delta$}} & \multicolumn{1}{c|}{} & \multicolumn{1}{c|}{} & \multicolumn{1}{c|}{} & \multicolumn{1}{c|}{} & \multicolumn{1}{c|}{} & \multicolumn{1}{c|}{} \\ \hline
\multicolumn{1}{|c|}{\textbf{0.4s}} & \multicolumn{1}{c|}{0.64} & \multicolumn{1}{c|}{0.46} & \multicolumn{1}{c|}{0.41} & \multicolumn{1}{c|}{1.52} & \multicolumn{1}{c|}{N/A} & \multicolumn{1}{c|}{N/A} \\ \hline
\multicolumn{1}{|c|}{\textbf{0.8s}} & \multicolumn{1}{c|}{0.84} & \multicolumn{1}{c|}{0.66} & \multicolumn{1}{c|}{0.59} & \multicolumn{1}{c|}{2.79} & \multicolumn{1}{c|}{N/A} & \multicolumn{1}{c|}{N/A} \\ \hline
\multicolumn{1}{|c|}{\textbf{1.2s}} & \multicolumn{1}{c|}{0.99} & \multicolumn{1}{c|}{0.79} & \multicolumn{1}{c|}{0.73} & \multicolumn{1}{c|}{3.92} & \multicolumn{1}{c|}{N/A} & \multicolumn{1}{c|}{N/A} \\ \hline
\multicolumn{1}{|c|}{\textbf{1.6s}} & \multicolumn{1}{c|}{1.14} & \multicolumn{1}{c|}{0.95} & \multicolumn{1}{c|}{0.83} & \multicolumn{1}{c|}{5.27} & \multicolumn{1}{c|}{N/A} & \multicolumn{1}{c|}{N/A} \\ \hline
\multicolumn{1}{|c|}{\textbf{2.0s}} & \multicolumn{1}{c|}{1.27} & \multicolumn{1}{c|}{1.02} & \multicolumn{1}{c|}{0.93} & \multicolumn{1}{c|}{6.36} & \multicolumn{1}{c|}{1.93} & \multicolumn{1}{c|}{1.31} \\ \hline
\end{tabular}

\begin{tabular}{ccccccc}
\multicolumn{1}{l}{} & \multicolumn{1}{l}{} & \multicolumn{1}{l}{} & \multicolumn{1}{l}{} & \multicolumn{1}{l}{} & \multicolumn{1}{l}{} & \multicolumn{1}{l}{} \\
\multicolumn{1}{l}{} & \multicolumn{6}{c}{\textbf{MAE\_X (Grids)}} \\ \cline{2-7} 
\multicolumn{1}{l|}{} & \multicolumn{1}{c|}{\multirow{2}{*}{\textbf{\begin{tabular}[c]{@{}c@{}}Proposed\\ ($\Omega=1$)\end{tabular}}}} & \multicolumn{1}{c|}{\multirow{2}{*}{\textbf{\begin{tabular}[c]{@{}c@{}}Proposed\\ ($\Omega=3$)\end{tabular}}}} & \multicolumn{1}{c|}{\multirow{2}{*}{\textbf{\begin{tabular}[c]{@{}c@{}}Proposed\\ ($\Omega=5$)\end{tabular}}}} & \multicolumn{1}{c|}{\multirow{2}{*}{\textbf{\begin{tabular}[c]{@{}c@{}}Baseline 1\\ (Kalman filter)\end{tabular}}}} & \multicolumn{1}{c|}{\multirow{2}{*}{\textbf{\begin{tabular}[c]{@{}c@{}}Baseline 2\\ (Basic LSTM)\end{tabular}}}} & \multicolumn{1}{c|}{\multirow{2}{*}{\textbf{\begin{tabular}[c]{@{}c@{}}Baseline 3\\ \cite{kim2017probabilistic}\end{tabular}}}} \\ \cline{1-1}
\multicolumn{1}{|c|}{\textbf{$\Delta$}} & \multicolumn{1}{c|}{} & \multicolumn{1}{c|}{} & \multicolumn{1}{c|}{} & \multicolumn{1}{c|}{} & \multicolumn{1}{c|}{} & \multicolumn{1}{c|}{} \\ \hline
\multicolumn{1}{|c|}{\textbf{0.4s}} & \multicolumn{1}{c|}{0.24} & \multicolumn{1}{c|}{0.15} & \multicolumn{1}{c|}{0.14} & \multicolumn{1}{c|}{0.54} & \multicolumn{1}{c|}{N/A} & \multicolumn{1}{c|}{N/A} \\ \hline
\multicolumn{1}{|c|}{\textbf{0.8s}} & \multicolumn{1}{c|}{0.30} & \multicolumn{1}{c|}{0.20} & \multicolumn{1}{c|}{0.18} & \multicolumn{1}{c|}{1.17} & \multicolumn{1}{c|}{N/A} & \multicolumn{1}{c|}{N/A} \\ \hline
\multicolumn{1}{|c|}{\textbf{1.2s}} & \multicolumn{1}{c|}{0.36} & \multicolumn{1}{c|}{0.25} & \multicolumn{1}{c|}{0.23} & \multicolumn{1}{c|}{1.38} & \multicolumn{1}{c|}{N/A} & \multicolumn{1}{c|}{N/A} \\ \hline
\multicolumn{1}{|c|}{\textbf{1.6s}} & \multicolumn{1}{c|}{0.43} & \multicolumn{1}{c|}{0.33} & \multicolumn{1}{c|}{0.27} & \multicolumn{1}{c|}{1.80} & \multicolumn{1}{c|}{N/A} & \multicolumn{1}{c|}{N/A} \\ \hline
\multicolumn{1}{|c|}{\textbf{2.0s}} & \multicolumn{1}{c|}{0.50} & \multicolumn{1}{c|}{0.35} & \multicolumn{1}{c|}{0.32} & \multicolumn{1}{c|}{2.07} & \multicolumn{1}{c|}{0.96} & \multicolumn{1}{c|}{0.44} \\ \hline
\end{tabular}

\begin{tabular}{ccccccc}
\multicolumn{1}{l}{} & \multicolumn{1}{l}{} & \multicolumn{1}{l}{} & \multicolumn{1}{l}{} & \multicolumn{1}{l}{} & \multicolumn{1}{l}{} & \multicolumn{1}{l}{} \\
\multicolumn{1}{l}{} & \multicolumn{6}{c}{\textbf{MAE\_Y (Grids)}} \\ \cline{2-7} 
\multicolumn{1}{l|}{} & \multicolumn{1}{c|}{\multirow{2}{*}{\textbf{\begin{tabular}[c]{@{}c@{}}Proposed\\ ($\Omega = 1$)\end{tabular}}}} & \multicolumn{1}{c|}{\multirow{2}{*}{\textbf{\begin{tabular}[c]{@{}c@{}}Proposed\\ ($\Omega = 3$)\end{tabular}}}} & \multicolumn{1}{c|}{\multirow{2}{*}{\textbf{\begin{tabular}[c]{@{}c@{}}Proposed\\ ($\Omega = 5$)\end{tabular}}}} & \multicolumn{1}{c|}{\multirow{2}{*}{\textbf{\begin{tabular}[c]{@{}c@{}}Baseline 1\\ (Kalman filter)\end{tabular}}}} & \multicolumn{1}{c|}{\multirow{2}{*}{\textbf{\begin{tabular}[c]{@{}c@{}}Baseline 2\\ (Basic LSTM)\end{tabular}}}} & \multicolumn{1}{c|}{\multirow{2}{*}{\textbf{\begin{tabular}[c]{@{}c@{}}Baseline 3\\ \cite{kim2017probabilistic}\end{tabular}}}} \\ \cline{1-1}
\multicolumn{1}{|c|}{\textbf{$\Delta$}} & \multicolumn{1}{c|}{} & \multicolumn{1}{c|}{} & \multicolumn{1}{c|}{} & \multicolumn{1}{c|}{} & \multicolumn{1}{c|}{} & \multicolumn{1}{c|}{} \\ \hline
\multicolumn{1}{|c|}{\textbf{0.4s}} & \multicolumn{1}{c|}{0.45} & \multicolumn{1}{c|}{0.34} & \multicolumn{1}{c|}{0.31} & \multicolumn{1}{c|}{1.29} & \multicolumn{1}{c|}{N/A} & \multicolumn{1}{c|}{N/A} \\ \hline
\multicolumn{1}{|c|}{\textbf{0.8s}} & \multicolumn{1}{c|}{0.63} & \multicolumn{1}{c|}{0.52} & \multicolumn{1}{c|}{0.46} & \multicolumn{1}{c|}{2.35} & \multicolumn{1}{c|}{N/A} & \multicolumn{1}{c|}{N/A} \\ \hline
\multicolumn{1}{|c|}{\textbf{1.2s}} & \multicolumn{1}{c|}{0.75} & \multicolumn{1}{c|}{0.63} & \multicolumn{1}{c|}{0.58} & \multicolumn{1}{c|}{3.51} & \multicolumn{1}{c|}{N/A} & \multicolumn{1}{c|}{N/A} \\ \hline
\multicolumn{1}{|c|}{\textbf{1.6s}} & \multicolumn{1}{c|}{0.87} & \multicolumn{1}{c|}{0.75} & \multicolumn{1}{c|}{0.67} & \multicolumn{1}{c|}{4.78} & \multicolumn{1}{c|}{N/A} & \multicolumn{1}{c|}{N/A} \\ \hline
\multicolumn{1}{|c|}{\textbf{2.0s}} & \multicolumn{1}{c|}{0.95} & \multicolumn{1}{c|}{0.81} & \multicolumn{1}{c|}{0.73} & \multicolumn{1}{c|}{5.84} & \multicolumn{1}{c|}{1.34} & \multicolumn{1}{c|}{1.06} \\ \hline
\end{tabular}
\end{table*}

\subsection{Data Collection}
  We collected the large set of vehicle trajectory data from several hours of highway driving around Seoul, Korea. The test vehicle was Hyundai Genesis equipped with 
  Delphi long range front radars (Fig.~\ref{fig:genesis}). 
  The sampling rate for the data was set to 10ms during the data collection, but the raw data was too noisy and often imbued with cut-offs because of asynchronous sampling. To cope with this problem, we averaged the samples over 100ms duration and thus the update period of the final data became 100ms. As a result, there were total 1325 trajectory sequences recorded from 26 different highway scenarios including lane changes, cut-in, and merging at the junction. 
  From the training dataset, we used 85\% (1126 sequences) for the training and 15\% (199 sequences) for validation.
\subsection{Experiment Setup}
  In our experiments, the maximum prediction range is set to $\Delta = 20$ (corresponding to 2 seconds). The initial learning rate is set to 0.0008 and halved whenever the validation error is plateaued. The mini batch size $B$ is set to 256. Through intensive empirical optimization, we determine the hyperparameters of the LSTM encoder-decoder network as
  \begin{itemize}
  \item  The depth of fully connected layers: 3
  \item  The LSTM cell state dimension: 256
  \item The depth of LSTM stack: 2
  \item The beam width $K$: 10
  \item The observation length $M$: 30
  \end{itemize} 
  Predicting trajectory sample every 0.1s requires $20$ decoder updates to reach the prediction horizon of 2 sec. We found that this is not desirable for the accuracy of long term prediction and it leads to high computational cost for inference. Hence, we trained the decoder to generate the trajectory sample every 0.2 sec. This allows the proposed system to reach the prediction horizon of 2 sec only with $10$ updates.
   
  For the performance evaluation, we use the Top-$\Omega$ mean absolute error (MAE) as a performance metric. Given the top $\Omega$ trajectory candidates among the trajectories produced by the proposed scheme, the Top-$\Omega$ MAE is obtained by finding the trajectory which is closest to the ground truth and evaluating the absolute error between them, i.e., 
       \begin{align}
  \mbox{Top-$\Omega$  MAE} = \frac{1}{L} \sum_{i=1}^{L} \left\| \begin{bmatrix}w_i^{*} \\ l_i^{*} \end{bmatrix} -   \begin{bmatrix}w_i^{(gt)} \\ l_i^{(gt)} \end{bmatrix}   \right\|_2
  \end{align}
where $L$ is the number of the test examples and $(w_i^{*}, l_i^{*})$ is the candidate grid index closest to the ground truth and $(w_i^{(gt)}, l_i^{(gt)})$ is the corresponding ground truth grid index, respectively.
   This metric explains how well the ground truth trajectory can be predicted by one of top $\Omega$ trajectory candidates. Note that the Top-$\Omega$ MAE is measured in the unit of the grid on the OGM. We can calculate the Top-$\Omega$ MAE for each future time step $\delta \in \{1,..., \Delta\}$, which is denoted as ``Top-$\Omega$ MAE$(\delta)$".
    We can measure the error in longitudinal and lateral directions using Top-$\Omega$ MAE\_X$(\delta)$ and Top-$\Omega$ MAE\_Y$(\delta)$, respectively.
\subsection{Experiment Results}

  We first present the MAE performance of the proposed prediction method. For performance comparison, we compare our method with the conventional Kalman filter based on the constant velocity model. We also consider the basic LSTM model trained to predict the probability of the occupancy over OGM and more complex LSTM based prediction method proposed in \cite{kim2017probabilistic}. We also present the performance of the proposed algorithm with $\Omega =1, 3$ and $5$. Table \ref{table:mae} presents the MAE, MAE\_X, and MAE\_Y achieved by the trajectory prediction algorithms of interest for the prediction time steps of $\Delta = 0.4, 0.8, 1.2, 1.6$ and $2.0$ sec. Since Kalman filter can also generate the future sequence by repeating the prediction update steps, we provide the MAEs for all time steps of $\Delta$. On the contrary, the LSTM model and the prediction scheme \cite{kim2017probabilistic} only predict the future location of the target vehicle after particular time ahead (e.g., $2.0$ sec). Hence, we provide the MAEs corresponding to $\Delta = 2.0$ sec only. We see that the prediction accuracy for all schemes of interest decreases with $\Delta$ since it is more difficult to predict the distant future. We observe that  the proposed algorithm with $\Omega = 1$ significantly outperforms both the Kalman filter and the basic LSTM model. With $\Omega = 1$, the proposed scheme achieves the performance comparable to the baseline \cite{kim2017probabilistic}. But with $\Omega = 3$ and $5$, the performance of the proposed scheme exceeds that of the baseline \cite{kim2017probabilistic} since our method generates more trajectory samples which are strongly likely to appear. Note that as compared to the baseline \cite{kim2017probabilistic}, the proposed scheme has the advantage of generating the full sequences of the promising trajectory candidates in each time step.   The above mentioned trend appears for all MAE, MAE\_X, and MAE\_Y results provided in Table~\ref{table:mae}.
  
\begin{figure}[t]
	\centering
    \includegraphics[width=\columnwidth]{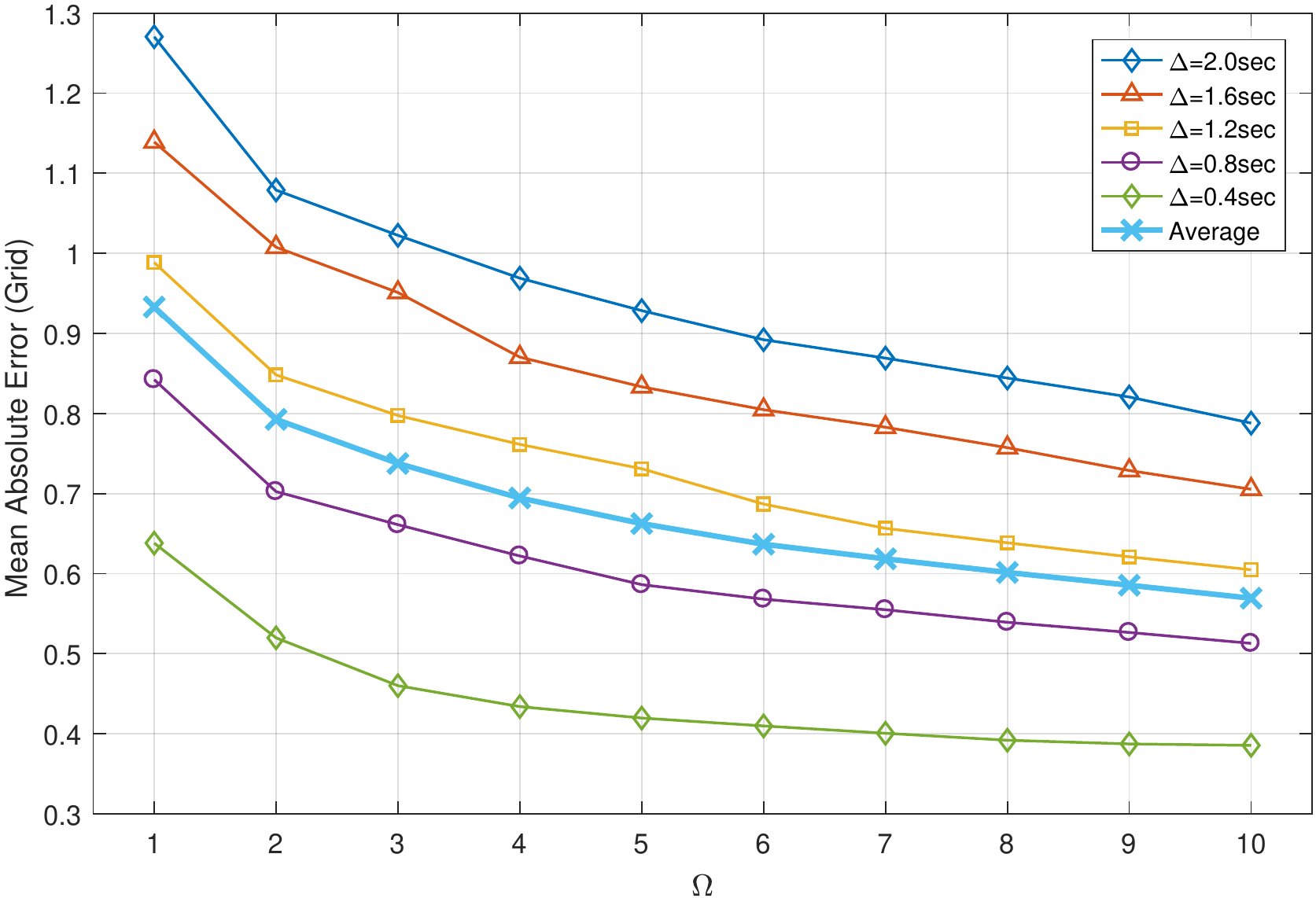}
    \caption{MAE versus $\Omega$ of the proposed method}
    \label{fig:maegraph}
\end{figure}

Next, we investigate how the proposed method behaves with respect to the parameters $\Delta$ and $\Omega$.  Fig.~\ref{fig:maegraph} shows the plot of MAE as a function of $\Omega$ for several values of $\Delta$. We also include the MAE averaged over all values of $\Delta$ considered. As $\Omega$ increases, the proposed system generates more trajectory hypotheses and the MAE performance improves for all cases. Note that the performance improvement due to increasing $\Omega$ is more dramatic for the case of higher $\Delta$,  yielding the decrease of MAE by around 0.2 grid for the change from $\Omega=1$ to $3$. This implies that including more trajectory hypotheses increases the probability that one of the trajectory candidates found by our algorithm is close to the ground truth trajectory. Therefore, the ability of the proposed method to generate the multiple trajectories would provide more informative and reliable prediction results to the subsequent path planning and control steps for autonomous driving.

%% file: tex/conclusion.tex
  In this paper, we proposed the new vehicle trajectory prediction method based on the LSTM encoder-decoder neural network architecture. The proposed system employs the LSTM encoder to analyze the past sensor measurements and the LSTM decoder to generate the future trajectory samples based on the encoder output. In order to alleviate the error propagation issue appearing in iterative decoding step, we applied the beam search algorithm, which keeps the $K$ best trajectory candidates for each decoding iteration. As a result, the proposed method can generate multiple trajectory hypotheses that might develop in many different ways given the same previous situation. Our experiment results confirmed that the proposed method can achieve the significant improvement over the existing methods in terms of the prediction accuracy while being able to generate the full sequence of the predicted trajectory in one shot.